\documentclass[sigconf,anonymous=false,review=false]{acmart}

\usepackage{makecell}    
\usepackage{algorithm}
\usepackage{algorithmic}
\usepackage{multirow}
\usepackage{enumitem}
\usepackage{caption}
\usepackage{booktabs}
\usepackage{amsmath,amsfonts}
\usepackage{textcomp}
\usepackage{subcaption}
\usepackage{footnote}
\usepackage{graphicx}
\usepackage{multirow}
\usepackage{graphicx}
\usepackage{enumitem}
  
\allowdisplaybreaks[4]
\usepackage{hyperref}

\theoremstyle{definition}

\newcommand\M{R\&F}

\AtBeginDocument{%
  \providecommand\BibTeX{{%
    \normalfont B\kern-0.5em{\scshape i\kern-0.25em b}\kern-0.8em\TeX}}}

\copyrightyear{2026}
\acmYear{2026}
\setcopyright{cc}
\setcctype{by}
\acmConference[SIGIR '26]{Proceedings of the 49th International ACM SIGIR Conference on Research and Development in Information Retrieval}{July 20--24, 2026}{Melbourne, VIC, Australia}
\acmBooktitle{Proceedings of the 49th International ACM SIGIR Conference on Research and Development in Information Retrieval (SIGIR '26), July 20--24, 2026, Melbourne, VIC, Australia}
\acmDOI{10.1145/3805712.3808583}
\acmISBN{979-8-4007-2599-9/2026/07}

\begin{document}

\title{R\&F-Inventory: A Large-Scale Dataset for Monotonic Inventory Estimation in Reach and Frequency Advertising}

\author{Yunshan	Peng}
\affiliation{
  \institution{Kuaishou Technology, Beijing, China}
  \city{}
  \country{}
}
\email{pengyunshan@kuaishou.com}

\author{Ji Wu}
\affiliation{
  \institution{Kuaishou Technology, Beijing, China}
  \city{}
  \country{}
}
\email{wuji03@kuaishou.com}

\author{Wentao Bai}
\affiliation{
  \institution{Kuaishou Technology, Beijing, China}
  \city{}
  \country{}
}
\email{baiwentao@kuaishou.com}

\author{Yunke Bai}
\affiliation{
  \institution{Kuaishou Technology, Beijing, China}
  \city{}
  \country{}
}
\email{baiyunke@kuaishou.com}

\author{Jinan Pang}
\affiliation{
  \institution{Kuaishou Technology, Beijing, China}
  \city{}
  \country{}
}
\email{pangjinan@kuaishou.com}

\author{Wenzheng Shu}
\affiliation{
  \institution{Kuaishou Technology, Beijing, China}
  \city{}
  \country{}
}
\email{shuwenzheng@kuaishou.com}

\author{Yanxiang Zeng}
\affiliation{
  \institution{Kuaishou Technology, Beijing, China}
  \city{}
  \country{}
}
\email{zengyanxiang@kuaishou.com}

\author{Xialong Liu}
\affiliation{
  \institution{Kuaishou Technology, Beijing, China}
  \city{}
  \country{}
}
\email{zhaolei16@kuaishou.com}

\author{Peng Jiang}
\authornote{Corresponding author.}
\affiliation{
  \institution{Kuaishou Technology, Beijing, China}
  \city{}
  \country{}
}
\email{jiangpeng@kuaishou.com}

\renewcommand{\shortauthors}{Yunshan Peng et al.}

\begin{abstract}
Reach and Frequency ({\M}) contract advertising is an important form of widely used brand advertising. Unlike performance advertising, {\M} contracts emphasize controllable delivery of UV and PV under given targeting, scheduling, and frequency control constraints. In practical systems, advertisers typically need to view the UV, PV change curves at different budget levels in real time when creating an  {\M} contract. However, most existing publicly available advertising datasets are based on independent samples, lacking a characterization of the core structure of the "budget-performance curve" (including UV and PV) in  {\M} contracts.This paper proposes and releases a large-scale {\M} contract inventory estimation dataset. This dataset uses the  {\M} contract context consisting of "targeting-scheduling-frequency control" as the basic context, providing observations of UV and PV corresponding to multiple budget points within the same context, thus forming a complete budget-performance curve. The dataset explicitly includes a time-window-based frequency control mechanism (e.g., "no more than 3 times within 5 days") and naturally satisfies the monotonicity and diminishing marginal returns characteristics in the budget and scheduling dimensions. We further derive the theoretical maximum exposure ceiling and use it as a consistency check to evaluate data quality and the feasibility of model predictions. Using this data set, this paper defines two standardized benchmark tasks: single-point performance prediction and reconstruction of budget-performance curves, and provides a set of reproducible baseline methods and evaluation protocols. This dataset can support systematic research on problems such as structural constraint learning, monotonic regression, curve consistency modeling, and  {\M} contract planning.The code for our experiments can be found at
\url{https://github.com/pengyunshan/RF-Inventory}.
\end{abstract}



\maketitle
\section{Introduction}
\begin{table*}[t]
  \caption{Comparison of advertising inventory forecasting resources and related systems.}
  \label{t:overall}
  \resizebox{\linewidth}{!}{
  \centering
  \begin{tabular}{lcccccc}
    \toprule
    \textbf{\makecell{Resource / Dataset}} & 
    \textbf{\makecell{ResourceCategory}} & 
    \textbf{\makecell{Publicly\\Available}} &
    \textbf{Monotonicity} & 
    \textbf{\makecell{Frequency\\Control}} & 
    \textbf{\makecell{Reproducible\\Benchmark}} \\
    \midrule
    Criteo Display Advertising~\cite{criteo2014display} 
      & Ad Log Dataset & $\checkmark$ & $\times$ & $\times$ & $\checkmark$ \\
    Avazu CTR~\cite{avazu2014ctr} 
      & Ad Log Dataset & $\checkmark$ & $\times$ & $\times$ & $\checkmark$ \\
    Alibaba / Taobao Ads~\cite{taobo2018ad,alibaba2020kddcupdebiasing} 
      & Ad Log Dataset & $\checkmark$ & $\times$ & $\times$ & $\checkmark$ \\
    \midrule
    Inventory / Reach Estimation (Literature)~\cite{agarwal2014budget,vaver2011predictive} 
      & Private Research Data & $\times$ & Partial & Partial & $\times$ \\
    \midrule
    Facebook Reach Planner~\cite{facebookreachplanner} 
      & Industrial Planning System & $\times$ & $\checkmark$ & $\checkmark$ & $\times$ \\
    Google Ads Reach Planner~\cite{googlereachplanner} 
      & Industrial Planning System & $\times$ & $\checkmark$ & $\checkmark$ & $\times$ \\
    \midrule
    Nielsen TV R\&F Measurement 
      & Media Measurement Data & Partial & $\checkmark$ & $\checkmark$ & $\times$ \\
    \midrule
    Demand / Supply Forecasting Benchmarks~\cite{m5forecasting} 
      & General Inventory Forecasting & $\checkmark$ & $\times$ & $\times$ & $\checkmark$ \\
    \midrule
    \textbf{Our Dataset (R\&F Inventory Dataset)} 
      & \textbf{R\&F Inventory Benchmark} & $\checkmark$ & $\checkmark$ & $\checkmark$ & $\checkmark$ \\
    \bottomrule
  \end{tabular}
  }
\end{table*}

\subsection{{\M} Contracts in Programmatic Advertising}

Reach and Frequency ({\M}) contracts are a prevalent and pivotal form of brand advertising in programmatic advertising systems, differing fundamentally from performance advertising that optimizes for clicks or conversions. Their core objectives are to maximize target audience reach within predefined budget and campaign constraints, while precisely regulating individual user ad exposure frequency \cite{criteo2014display}. Throughout this paper, we use \textbf{PV} (page views / impressions) and \textbf{UV} (unique visitors / covered people) as the two primary delivery outcome metrics, and treat the paired terms as synonymous. When creating an {\M} order, advertisers specify targeting criteria (e.g., region, age, gender), campaign delivery schedules, and time-window-based frequency control rules (e.g., "no more than 3 exposures per user within 5 days"). Based on these configurations, the advertising system is required to generate budget-performance curves—the core output for {\M} planning—that project key metrics (PV, UV, average exposure frequency) against varying budget levels.
This curve is critical for both advertisers and platforms: it guides advertisers'budget decisions and campaign expectation setting, and serves as a foundational basis for platforms' inventory planning and resource allocation. An ideal budget-performance curve must satisfy three key properties: monotonic increase with budget growth, diminishing marginal returns, and strict compliance with physical feasibility constraints imposed by scheduling and frequency control rules. However, constructing such a numerically reasonable and structurally consistent curve remains a major challenge in real-world advertising systems.

\subsection{Existing Research and Data Resources}
Mainstream advertising machine learning research focuses on click-through rate (CTR) and conversion rate (CVR) prediction, with evaluations relying on public advertising log datasets. These datasets take single PV or click events as basic sample units, focusing solely on point-level prediction and lacking explicit modeling of {\M} core attributes (budgets, schedules, frequency control). As such, they cannot directly support inventory estimation and budget-performance curve modeling for {\M} contracts.
While industrial systems (e.g., Reach Planners) can generate {\M} performance estimates, their underlying models and data are closed-source, precluding systematic academic analysis and experimental reproduction and creating a significant gap between industrial practice and academic research. Public data resources further exacerbate this issue: existing advertising datasets lack budget and frequency control dimensions, while general inventory forecasting datasets focus on product sales/supply and fail to capture the unique audience accessibility and frequency control constraints of advertising inventory. This not only hinders unified evaluation of {\M}-oriented modeling methods (e.g., structural constraint learning for budget-performance curves) but also severely limits the comparability of relevant research works.
Additionally, {\M} inventory estimation has inherent structural characteristics: for fixed configurations, UV and PV across budget levels are not independent, but form a curve bound by non-negotiable structural constraints (e.g., theoretical maximum PV determined by scheduling and frequency control). Ignoring these constraints can render model outputs practically unusable in real systems—even with minor point-level prediction errors—undermining advertiser experience and platform operational decisions.

\subsection{Contributions: The {\M} Dataset}
To address the aforementioned research and data gaps, this paper proposes and releases a structured dataset for {\M} contract inventory forecasting. The dataset takes {\M} contract contexts (targeting-scheduling-frequency control triads) as basic units, and provides UV/PV observations for multiple budget points within the same context—enabling the explicit construction of complete budget-performance curves. Notably, it includes detailed time-window-based frequency control rules, allowing the derivation of theoretical maximum exposure ceilings for model consistency verification.
Based on this dataset, we define two standardized research tasks for the {\M} scenario: single-point performance prediction and budget-performance curve modeling. We also design a unified data partitioning strategy and evaluation protocol to support systematic research on structured constraint learning and inventory prediction methods. Ultimately, this dataset aims to serve as an open, reproducible benchmark platform for {\M} contract inventory prediction, budget-performance modeling, and related structured learning problems—bridging the academia-industry gap and advancing research in this under-explored field.

\section{Related Work}
In this section, we will introduce related work on this dataset, including tasks related to advertising effectiveness prediction and inventory forecasting.

\subsection{Advertising Datasets and Inventory Estimation}
Publicly available advertising datasets have played a critical role in the development of advertising machine learning models, especially for performance prediction tasks such as click-through rate (CTR) and conversion rate (CVR) estimation. Representative examples include the Criteo Display Advertising Dataset \cite{criteo2014display}, the Avazu CTR dataset \cite{avazu2014ctr}, and large-scale e-commerce advertising logs released by Alibaba and Taobao \cite{taobo2018ad,alibaba2020kddcupdebiasing}. These datasets typically treat impressions, clicks, or conversions as independent samples, with rich user, ad, and contextual features, and have become standard benchmarks for supervised learning in performance advertising.

However, such datasets are primarily designed for pointwise prediction tasks and do not explicitly model budget conditions, campaign scheduling, or frequency control constraints, which are fundamental components in brand advertising and Reach and Frequency (R\&F) contract delivery. In these datasets, budget information—if present at all—is usually implicit or fixed, and performance outcomes at different budget levels are not jointly modeled. As a result, they are ill-suited for inventory estimation, coverage forecasting, or budget–performance curve modeling, where observations under different budgets within the same campaign context are structurally dependent rather than independent.

In contrast, R\&F contract planning requires advertisers to reason about how UV and PV evolve as the budget or delivery horizon increases, subject to scheduling and frequency constraints. The resulting outcomes naturally form structured curves with monotonicity, saturation, and diminishing marginal returns. This curve-level structure is not systematically captured in existing public advertising datasets, limiting their applicability to R\&F inventory forecasting and planning tasks.

In industrial systems, major advertising platforms provide internal tools—commonly referred to as Reach and Frequency Planners—to forecast reach and exposure under given targeting, scheduling, and frequency control settings \cite{facebookreachplanner,googlereachplanner}. These planners play a central role in campaign decision-making and implicitly model inventory availability and delivery feasibility. However, the underlying data and models are proprietary, and the outputs are typically provided only through closed interfaces, making them unsuitable as reproducible benchmarks for academic research. While some prior studies analyze advertising reach or inventory forecasting using private data \cite{vaver2011predictive}, the lack of public, structured datasets has hindered systematic comparison and reproducibility.

Recent efforts such as the Meta Ad Library \cite{meta2026adlibrary} improve transparency by releasing aggregate spend and PV ranges for running advertisements, but they are not designed to support controlled inventory estimation or curve modeling under fixed R\&F contract contexts.

In contrast, the dataset proposed in this paper adopts the R\&F contract context—defined by targeting, scheduling, and frequency control—as the basic unit, and provides multiple real-world observations under different budget levels within the same context. By explicitly incorporating scheduling and time-window-based frequency control information, our dataset offers a public and reproducible foundation for inventory and reach estimation under realistic budget constraints.

\subsection{Budget--Performance Modeling}
\label{subsec:dm}
Modeling the relationship between advertising budget and delivery outcomes has long been studied in the context of budget allocation, bidding optimization, and campaign planning. Classical and modern works commonly assume that advertising effectiveness increases monotonically with budget while exhibiting diminishing marginal returns \cite{borgs2007dynamics,agarwal2014budget,yuan2013realtime}. These assumptions are widely used to design budget optimization and pacing algorithms in sponsored search and display advertising systems.

While these studies provide important insights into the economic and algorithmic properties of budget–performance relationships, their empirical evaluations often rely on proprietary logs or synthetic simulations, and do not offer publicly available benchmarks for reproducible evaluation. Moreover, most prior work focuses on optimization given a known response function, rather than learning the response curve itself from structured observations under realistic delivery constraints.

From a machine learning perspective, budget–performance curve modeling is closely related to structural constraint learning, particularly regression problems with monotonicity or shape constraints. Foundational work on isotonic regression and ordered inference provides theoretical tools for enforcing monotonic relationships \cite{bartholomew1972isotonic}. More recent approaches introduce flexible models with explicit monotonic constraints, including calibrated lattice models \cite{gupta2016monotonic} and deep neural architectures supporting partial monotonicity \cite{you2017deep}. These methods have been shown to improve interpretability, stability, and reliability in domains where physical or economic constraints are known a priori.

Despite these methodological advances, their systematic evaluation in advertising inventory and reach estimation remains limited by data availability. Existing public datasets lack a budget dimension or curve-level supervision, preventing meaningful assessment of models that aim to capture monotonicity, saturation, or feasibility constraints. Furthermore, practical constraints introduced by campaign scheduling and time-window-based frequency control are often treated as implicit assumptions, rather than being incorporated into explicit modeling or evaluation protocols.

The dataset introduced in this paper addresses these limitations by providing structured budget–performance curves and explicitly modeling scheduling and frequency control mechanisms. This enables budget–performance modeling to be formulated as a learning problem with explicit structural constraints, and supports curve-level evaluation beyond pointwise accuracy. As such, our dataset establishes a unified and reproducible benchmark for research on structured constraint learning, monotonic modeling, and inventory forecasting in R\&F contract scenarios.

\section{Data Description}
\begin{figure}[t]
  \includegraphics[width=0.5\textwidth]{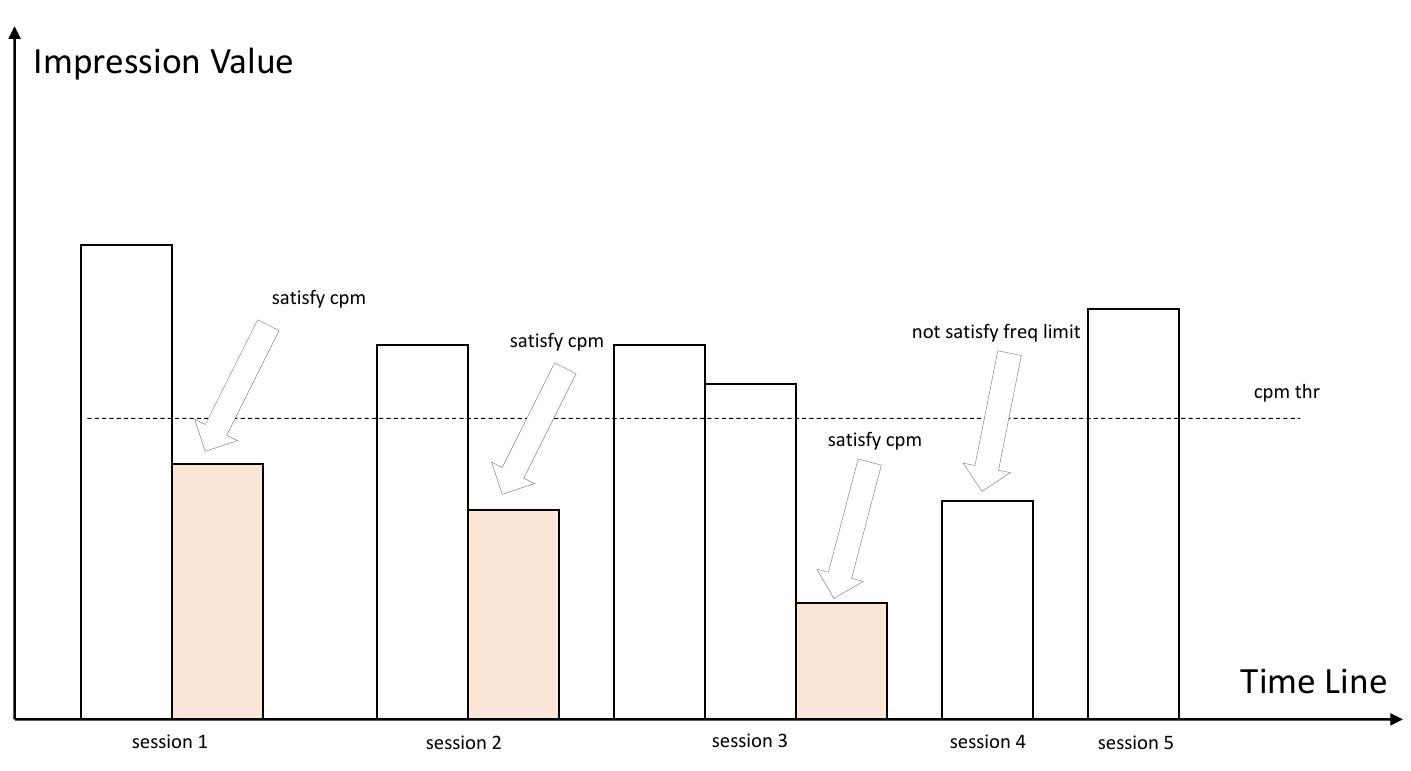}
  \caption[width=0.5\textwidth]{Simulation diagram of a single user inserting a CPM ad}
  \label{fig:framework}
\end{figure}

\subsection{Data Collection}
\label{subsec:blend}
In the dataset constructed for this study, an  {\M} contract context is defined as a delivery configuration determined jointly by three categories of information, which are specified as follows:
Targeting criteria: Screening constraints for the target audience, such as region, age group, and gender;
Delivery schedule: The start time, end time of ad delivery, and the corresponding actual delivery days;
Frequency control rules: Upper limits of ad exposure frequency based on time windows. For example, a cap of 3 exposures within 7 days means that the number of exposures to the same ad for a single user must be no more than 3 in any 7-day time window.
After fixing a specific  {\M} contract context, the UV and PV corresponding to different budget levels form a set of budget-performance observation points. Sorting these observation points along the budget dimension yields a budget-performance curve. Unlike the approach in existing studies that treats each budget point as an independent sample, this study explicitly preserves the intrinsic structural relationships among budget-performance observation points under the same  {\M} contract context.
Taking the scenario with fixed targeting criteria, a 10-day delivery schedule and a frequency control rule of no more than 3 exposures within 7 days as an example, its budget-performance relationship can be characterized by a two-dimensional curve: the horizontal axis of the curve represents different CPM (Cost Per Mille) thresholds, the vertical axis denotes the UV and PV under the corresponding CPM thresholds, and the delivery budget can be calculated as the product of PV and the CPM threshold.
This dataset is constructed based on user sampling, and the relevant data on PV and UV achievable through ad delivery under different combinations of frequency control rules, delivery schedules and CPM thresholds is generated via an offline simulation

For each user(as illustrated in Figure \ref{fig:framework}), we construct a chronological sequence of ad exposures within a given time window, where each ad request (identified by a unique session ID,) may return one or two ads.

Given a candidate CPM threshold, we evaluate whether a new Reach and Frequency (R\&F) ad can be inserted into the exposure sequence by replacing an existing non-R\&F ad whose actual CPM is lower than the given threshold, while still satisfying the predefined frequency control constraints.

If such an insertion is feasible, we count this user as contributing one additional PV and one additional UV to the R\&F contract under the corresponding CPM threshold.

The resulting inventory signals reflect feedback from a real advertising system, while no personally identifiable user information is exposed or released.
\subsection{Statistics and Usage}
\label{subsec:egcd}
Each record in the dataset can be represented as the following 6-tuple: <x, T, fc, b, R, I>

Among them, x denotes the targeting feature, T represents the length of the delivery schedule (unit: day), and fc=(W,K) stands for the frequency control rule (W is the length of the frequency control window, K is the upper limit of exposure frequency within the corresponding window). b refers to the delivery budget, R is UV, and I is PV. b and I can uniquely determine a budget-performance observation point, and the final CPM (Cost Per Mille) can be calculated by the formula CPM = 1000*b/I. To ensure that all data can be statistically analyzed in the same dimension, we set a series of different CPM thresholds, ranging from 0.4 to 3.0 with a uniform interval of 0.05, resulting in a total of 53 threshold points selected.

At the data organization level, we further grouped the records with the same (x, T, fc) combination, and each group corresponds to one budget-performance curve. On the premise of fixing (x, fc), the longer the delivery schedule, the larger the PV (I) and UV (R); meanwhile, the higher the budget, the greater the PV (I) and UV (R). This law is consistent with the basic common sense of advertising delivery.

Each record in the dataset can be represented as the following 6-tuple: <x, T, fc, b, R, I>

The dataset adopts a time window-based frequency control mechanism, which restricts that the number of exposures for any user within any consecutive W days shall not exceed K. This frequency control mechanism has a significant interaction with the delivery schedule: under fixed frequency control parameters, the longer the delivery cycle, the higher the maximum cumulative number of exposures that a single user can achieve throughout the schedule, thereby increasing the overall upper limit of achievable exposures.

Given the delivery schedule length T, frequency control parameters (W,K), and UV R, there exists an upper bound for the maximum number of exposures of a single user during the entire delivery cycle, which is expressed as follows:
\begin{align}
    m_{max}(T;W,K)=K\cdot \lceil \frac{T}{W} \rceil
\end{align}

Based on this, the theoretical maximum total number of exposures of the  {\M} contract under given conditions can be derived:

\begin{align}
    I_{max}=R\cdot K \cdot \lceil \frac{T}{W}\rceil
\end{align}
In this study, this upper bound is used to perform consistency checks on the samples in the dataset, and it is taken as a feasible region constraint during the model prediction stage.

At the data organization level, we further grouped the records with the same (x, T, fc) combination, and each group corresponds to one budget-performance curve. On the premise of fixing (x, fc), the longer the delivery schedule, the larger the PV (I) and UV (R); meanwhile, the higher the budget, the greater the PV (I) and UV (R). This law is consistent with the basic common sense of advertising delivery.

\section{Potential Research Directions}
\subsection{Two Type of Tasks}
\label{subsec:task}

Centering on this dataset and combining practical business problems, we define the following two standardized benchmark tasks. These two tasks range from simple to complex, which is more in line with actual application requirements.

\noindent \textbf{Task 1: Single-Point Performance Prediction.} In this task, given the conditions of <targeting, delivery schedule, frequency control, CPM threshold>, the model is required to predict the corresponding UV and PV. This task corresponds to the performance prediction requirement in actual systems when advertisers input a single budget.

In practical business scenarios, when an advertiser provides a specific budget, we need to output the corresponding UV and PV. This problem has practical business significance: overestimating UV and PV will cause losses to the platform, while underestimating them will lead to losses to advertisers. Therefore, the accuracy of prediction is of great importance.

\noindent \textbf{Task 2: Multi-Dimensional Monotonic Delivery Outcome Estimation.} In Reach and Frequency ({\M}) contract planning, advertisers and systems focus not only on the delivery outcome under a specific conditional point but also on the overall change trend of delivery outcomes when the budget, delivery schedule, or bid threshold changes. In actual systems, delivery outcomes usually show stable structural rules with the increase of delivery intensity and duration: under the same other conditions, a longer delivery schedule will not reduce the deliverable PV and UV; a higher budget or CPM threshold will not decrease the delivery opportunities available to the system; and looser frequency control constraints will not lead to a decline in delivery outcomes.

Task 2 formulates inventory estimation as a supervised learning problem with multi-dimensional monotonic constraints. Given delivery context features, the model predicts delivery outcomes at a single conditional point, while the evaluation examines prediction accuracy and monotonic consistency across scheduling, and CPM dimensions within the same context.

The task defines each sample as consisting of the following two parts:
\begin{itemize}[leftmargin=*]
    \item \textbf{Context}: Targeting criteria x and frequency control window W (as well as other non-monotonic constraint fields);
    \item \textbf{Monotone Inputs}: Delivery schedule length T, CPM threshold indicating the set of delivery opportunities available to the system under this threshold.
\end{itemize}

The goal of the model is to learn the mapping relationship:

\begin{align}
    (x, W, fc;\; T, B) \;\longrightarrow\; (\hat R, \hat I)
\end{align}

where $\hat R$ and $\hat I$ represent the predicted UV and PV, respectively.

In the dataset provided in this study, delivery intensity is mainly characterized by the CPM grid (e.g., discrete values in [0.4, 4.0]). Therefore, in practical use, Task 2 can also be simplified as: 
\begin{align}
    (x, W, fc;\; T, \mathrm{cpm}) \;\longrightarrow\; (\hat R, \hat I)
\end{align}

where the budget can be used as a derived quantity for analysis or consistency checks, rather than being taken as an explicit input dimension.

For the same context (i.e., the same targeting criteria and non-monotonic fields), we introduce a partial order relationship in the delivery condition space. Let $\mathbf{u}_1=(T_1,B_1)$ and $\mathbf{u}_2=(T_2,B_2)$; if the following conditions are satisfied:

\begin{align}
    T_1 \le T_2,\quad B_1 \le B_2
\end{align}

then the model predictions are required to satisfy: 

\begin{align}
    \hat R(\mathbf{u}_1) \le \hat R(\mathbf{u}_2),\quad \hat I(\mathbf{u}_1) \le \hat I(\mathbf{u}_2)
\end{align}

This constraint belongs to partial-order monotonicity: the prediction results are only required to be non-decreasing when all monotonic dimensions are non-decreasing (and at least one dimension is strictly increasing). This setting avoids imposing unnecessary constraints on incomparable conditional points and is consistent with the actual  {\M} planning logic.

\subsection{Dataset Splitting Methods}
\begin{figure}[t]
  \includegraphics[width=0.5\textwidth]{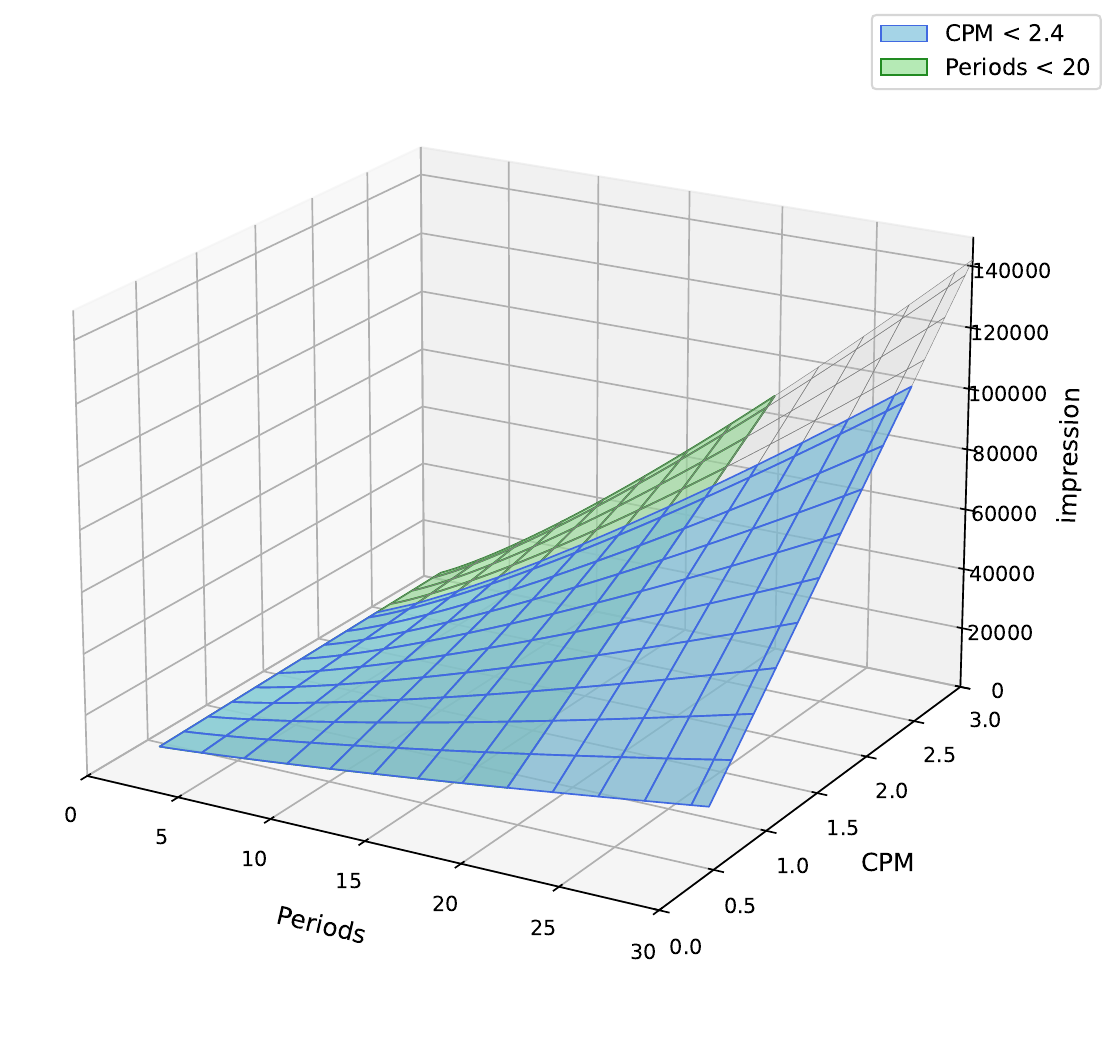}
  \caption[width=0.5\textwidth]{Schematic diagram of dataset splitting methods}
  \label{fig:train_test}
\end{figure}
\label{subsec:datasplit}

To evaluate the generalization ability of the model under different delivery conditions, this study does not adopt random sample-level splitting. Instead, based on two key control variables—delivery intensity and delivery duration—we design three dataset splitting methods with clear business implications, as illustrated in Figure \ref{fig:train_test}. All splitting operations are performed under the premise of fixed targeting criteria and frequency control windows to avoid simple data leakage.

\subsubsection{Splitting Method 1: CPM-based Extrapolation Split}
\label{subsubsec:method1}

This splitting method divides the samples in the dataset based on the CPM threshold, with specific rules as follows:

Training set: Samples with CPM thresholds lower than a given standard;

Test set: Samples with CPM thresholds higher than the given standard.

The core purpose of this splitting method is to evaluate the model's ability to predict delivery outcomes under higher CPM thresholds after being trained under low delivery intensity conditions. This setting corresponds to a typical application scenario in practical business: "inferring the possible effects of more aggressive delivery strategies based on the results of conservative bidding or low-budget test delivery".

\subsubsection{Splitting Method 2: Schedule-based Extrapolation Split}
\label{subsubsec:method2}

This splitting method classifies samples based on the length of the delivery schedule, with specific rules as follows:

Training set: Samples with shorter delivery schedules;

Test set: Samples with longer delivery schedules.

This setting is mainly used to evaluate whether the model can learn the inherent law of delivery outcome growth over time from short-term delivery data, and reasonably predict the scale of UV and PV under longer delivery cycles. This splitting method directly corresponds to the common business process of "short-term evaluation and long-term planning" in  {\M} contract planning.

\subsubsection{Splitting Method 3: Joint CPM–Schedule Extrapolation Split}
\label{subsubsec:method3}

This splitting method classifies samples that simultaneously meet the two conditions of high CPM threshold and long delivery schedule as the test set, and all other samples as the training set. Formally, the test set consists of samples that satisfy the following conditions:
\begin{align}
    \mathrm{cpm} \ge \tau_{\mathrm{cpm}} \;\wedge\; T \ge \tau_T
\end{align}

This splitting method is used to evaluate the model's generalization ability in the most challenging scenarios, i.e., the model can only observe relatively conservative delivery conditions (low CPM, short schedule) during the training phase, but needs to simultaneously predict delivery outcomes under high delivery intensity and long delivery cycles during the test phase. This setting is consistent with the actual requirement for model robustness during the "large-scale delivery" phase in  {\M} contracts.

\subsection{Evaluation Metrics}
\label{subsec:eval}
The evaluation of Task 2 consists of two parts: point-level prediction accuracy and structural consistency evaluation. The model outputs predictions at each conditional point, and the evaluation is aggregated at the context level to ensure the comprehensiveness and rationality of the evaluation results.

\noindent \textbf{Point-Level Prediction Error}.On all conditional points of the test set, we calculate the following four error metrics to measure the prediction accuracy of UV and PV respectively:
\begin{itemize}[leftmargin=*]

\item Mean Absolute Error (MAE) and Root Mean Square Error (RMSE) for UV (R);

\item Mean Absolute Error (MAE) and Root Mean Square Error (RMSE) for PV (I).
\end{itemize}

To avoid the dominant influence of large-scale contexts on the overall evaluation results, in addition to calculating the overall error of the entire test set, this study also reports the error metrics averaged by context to more objectively reflect the model's prediction performance under different contexts.

\noindent \textbf{Monotonic Consistency Evaluation}.For each context, we first construct a set of conditional point pairs that satisfy the partial order relationship defined earlier, which is formally expressed as:

\begin{align}
    \mathcal{P}=\{(\mathbf{u}_1,\mathbf{u}_2)\mid \mathbf{u}_1 \preceq \mathbf{u}_2\}
\end{align}

On this set of point pairs, we evaluate the monotonic consistency of the model's prediction results from two dimensions, with specific metrics as follows:

\begin{itemize}[leftmargin=*]

\item Violation Rate: The proportion of point pairs that violate the multi-dimensional monotonic constraints among all point pairs satisfying the partial order relationship, which is used to characterize the occurrence frequency of monotonic violations;

\item Violation Magnitude: The average value of the decline range of prediction results (UV or PV) among all point pairs that violate the monotonic constraints, which is used to measure the severity of monotonic violations.

\end{itemize}
The above two metrics are both applicable to the monotonic consistency evaluation of UV (R) and PV (I), which can comprehensively reflect the degree to which the model adheres to the multi-dimensional monotonic constraints during the prediction process.

\subsection{Experimental Results}
\label{subsec:result}



We evaluate a set of reproducible baseline methods for Task 2 under the joint extrapolation splitting setting (Method 3), where contexts with higher CPM thresholds and longer delivery schedules are held out for testing. The baselines include a conventional tree-based model (GBDT~\cite{friedman2001greedy}) and several models explicitly designed to enforce monotonicity, including POSNN~\cite{you2017deep}, SMM~\cite{igel2023smooth}, MN~\cite{sill1997monotonic}, PWL~\cite{gupta2019incorporate}, and a Hint-based model~\cite{sill1996monotonicity}.

As shown in Table~\ref{t:baseline}, models with explicit monotonic constraints consistently outperform unconstrained or weakly constrained models in both predictive accuracy and structural consistency. In particular, PWL, SMM, and MN achieve substantially lower MAE and RMSE for both impressions (PV) and reach (UV), while maintaining near-zero monotonicity violation rates.

Although the dataset provides a theoretically derived upper bound on maximum feasible exposure, we do not include it as an explicit evaluation metric, as none of the evaluated baselines explicitly model or meaningfully differentiate predictions near this saturation boundary.

\begin{table}[t]
  \caption{Performance comparison between baseline methods. 
  The best performance in each column is shown in \textbf{bold}}
  \label{t:baseline}
  \resizebox{\linewidth}{!}{
  \centering
  \begin{tabular}{lccccccc}
    \toprule
    \textbf{Method} & \textbf{Split} &PV MAE &PV RMSE & UV MAE & UV RMSE 
    & \makecell{Violation \\ PV \\ Rate} & \makecell{Violation \\ UV \\ Rate} \\
    \midrule
    GBDT~\cite{friedman2001greedy}  
      & Method 3
      & 442453 & 1044529 & 779 & 1312 & 0.61 & 0.56 \\
    \midrule
    POSNN~\cite{you2017deep}  
      & Method 3
      & 15314 & 24527 & 430 & 614 & \textbf{0.0} & \textbf{0.0} \\
    \midrule
    SMM~\cite{igel2023smooth}  
      & Method 3
      & 11841 & 21946 & 344 & 420 & \textbf{0.0} & \textbf{0.0} \\
    \midrule
    MN~\cite{sill1997monotonic}  
      &  Method 3
      & 11979 & 22989 & \textbf{260} & \textbf{338} & \textbf{0.0} & \textbf{0.0} \\
    \midrule
    PWL~\cite{gupta2019incorporate}  
      &  Method 3
      & \textbf{10298} & \textbf{21669} & 301 & 386 & \textbf{0.0} & \textbf{0.0} \\
    \midrule
    Hint~\cite{sill1996monotonicity}  
      &  Method 3
      & 16227 & 39577 & 444 & 699 & 0.31 & 0.11 \\
    \bottomrule
  \end{tabular}
  }
\end{table}

\section{Conclusion and Limitations}
This paper proposes and publishes a structured dataset for contract inventory forecasting, which explicitly models the interrelationships between budget, scheduling, and frequency control. By defining standardized learning tasks and evaluation protocols, we hope this dataset will facilitate systematic research on the problem of structured consistency advertising forecasting. Moreover, the dataset captures realistic delivery constraints such as monotonicity and diminishing returns, providing a practical and physically grounded benchmark for evaluating model behavior beyond point-wise accuracy.However, the analysis and baseline of this dataset still have some shortcomings and room for optimization. For example, the relationship between UV and PV is not explored; PV and UV are estimated separately, and the provided baseline method does not share underlying parameters. The monotonic relationship between the frequency control window and its quantity is also not explored and is treated as a fixed feature in our paper. 

\balance
\bibliographystyle{ACM-Reference-Format}
\bibliography{reference}

\end{document}